\documentclass{article}
\usepackage{ijcai22}

\usepackage{times}
\usepackage{soul}
\usepackage{url}
\usepackage[hidelinks]{hyperref}
\usepackage{natbib}
\usepackage[utf8]{inputenc}
\usepackage[small]{caption}
\usepackage{graphicx}
\usepackage{amsmath}
\usepackage{amsthm}
\usepackage{latexsym}
\usepackage{spverbatim}

\usepackage{amsfonts}       

\usepackage{nicefrac}       
\usepackage{microtype}      
\usepackage{hhline}
\usepackage{boxedminipage}
\usepackage{hyperref}
\usepackage{amssymb}
\usepackage{wrapfig,lipsum,booktabs}
\usepackage{cleveref}
\usepackage{booktabs}
\usepackage{bm}
\usepackage{algorithm}
\usepackage{algpseudocode}
\usepackage{makecell}
\usepackage{graphicx}
\usepackage{nccmath}
\usepackage{caption}
\usepackage{multirow}
\usepackage{makecell}
\usepackage{subcaption}
\usepackage{tabularx}
\usepackage{colortbl}

\usepackage{microtype}
\PassOptionsToPackage{hyphens}{url}\usepackage{hyperref}

\newcommand{\B}[1] {\boldsymbol{#1}}
\def\bA{{\B{A}}}

\def\bH{{\B{H}}}

\def\bQ{{\B{Q}}}
\def\bK{{\B{K}}}

\def\bV{{\B{V}}}
\def\bX{{\B{X}}}
\def\bW{{\B{W}}}

\def\bv{{\B{v}}}

\title{Human Parity on CommonsenseQA: Augmenting Self-Attention with External Attention}

\makeatletter
\newcommand{\thickhline}{%
	\noalign {\ifnum 0=`}\fi \hrule height 1pt
	\futurelet \reserved@a \@xhline
}
\makeatother

\author{Yichong Xu, Chenguang Zhu, Shuohang Wang, Siqi Sun, Hao Cheng, \\
	 Xiaodong Liu, Jianfeng Gao, Pengcheng He, Michael Zeng \And Xuedong Huang 
	 \affiliations
Microsoft Corporation
\emails
   \{yicxu,chezhu,shuowa,siqi.sun,chehao,xiaodl,jfgao,penhe,nzeng,xdh\}@microsoft.com \\
}

\begin{document}

\maketitle

\begin{abstract}
 Most of today's AI systems focus on using self-attention mechanisms and transformer architectures on large amounts of diverse data to achieve impressive performance gains. In this paper, we propose to augment the transformer architecture with an external attention mechanism to bring external knowledge and context to bear.
 By integrating external information into the prediction process, we hope to reduce the need for ever-larger models and increase the democratization of AI systems.
 We find that the proposed external attention mechanism can significantly improve the performance of existing AI systems, allowing practitioners to easily customize foundation AI models to many diverse downstream applications. 
 In particular, we focus on the task of Commonsense Reasoning, demonstrating that the proposed external attention mechanism can augment existing transformer models and significantly improve the model's reasoning capabilities. The proposed system, Knowledgeable External Attention for commonsense Reasoning (KEAR), reaches human parity on the open CommonsenseQA research benchmark with an accuracy of 89.4\% in comparison to the human accuracy of 88.9\%. 
\end{abstract}
\section{Introduction}
\label{sec:intro}
Transformers~\citep{transformer} have revolutionized many areas of AI with state-of-the-art performance in a wide range of tasks~\citep{bert,visiontransformer}. The most notable and effective component in a Transformer model is the self-attention mechanism, which enables the model to dynamically leverage different parts of the input for computation, with no information loss for even the most distant parts of the input. With the success of pre-trained models~\citep{bert,roberta}, the Transformer and its self-attention mechanism have been widely adopted as the cornerstone of foundation models trained on huge amounts of data~\citep{foundationmodel}. 

One phenomenon found during the development of Transformer models is that models with larger sizes tend to have better learning abilities, especially when combined with large-scale data.
This has prompted the recent boom of super large Transformer models, ranging from BERT~\citep{bert} with 110 million parameters, to GPT-3~\citep{gpt3} with 175 billion parameters. 
Nevertheless, numerous studies have shown that the corresponding understanding and generation capabilities of these huge models are still behind humans  \citep{foundationmodel}. Furthermore, the sheer size of these models already poses serious practical challenges in utilization, deployment, interpretation, and environmental impact~\citep{carbon}. Thus, the recent ``scaling-up'' approach to Transformer-based NLP modeling is unsustainable and has been questioned in recent studies~\citep{foundationmodel}.

In this paper, we take a step back and examine the mechanism of current Transformer-based models. Self-attention was designed to allow the model to better analyze the inner structure of input data, and the model is trained to have its parameters grasp and memorize all the content and patterns of the training data. When the model is given a novel input $X$, the implicitly stored knowledge in the parameters about related information is activated to facilitate the analysis of $X$. This could partly explain why larger models pre-trained with more data have an advantage in performance.

\begin{figure*}[!h]
	\centering
	\includegraphics[width=0.92\linewidth]{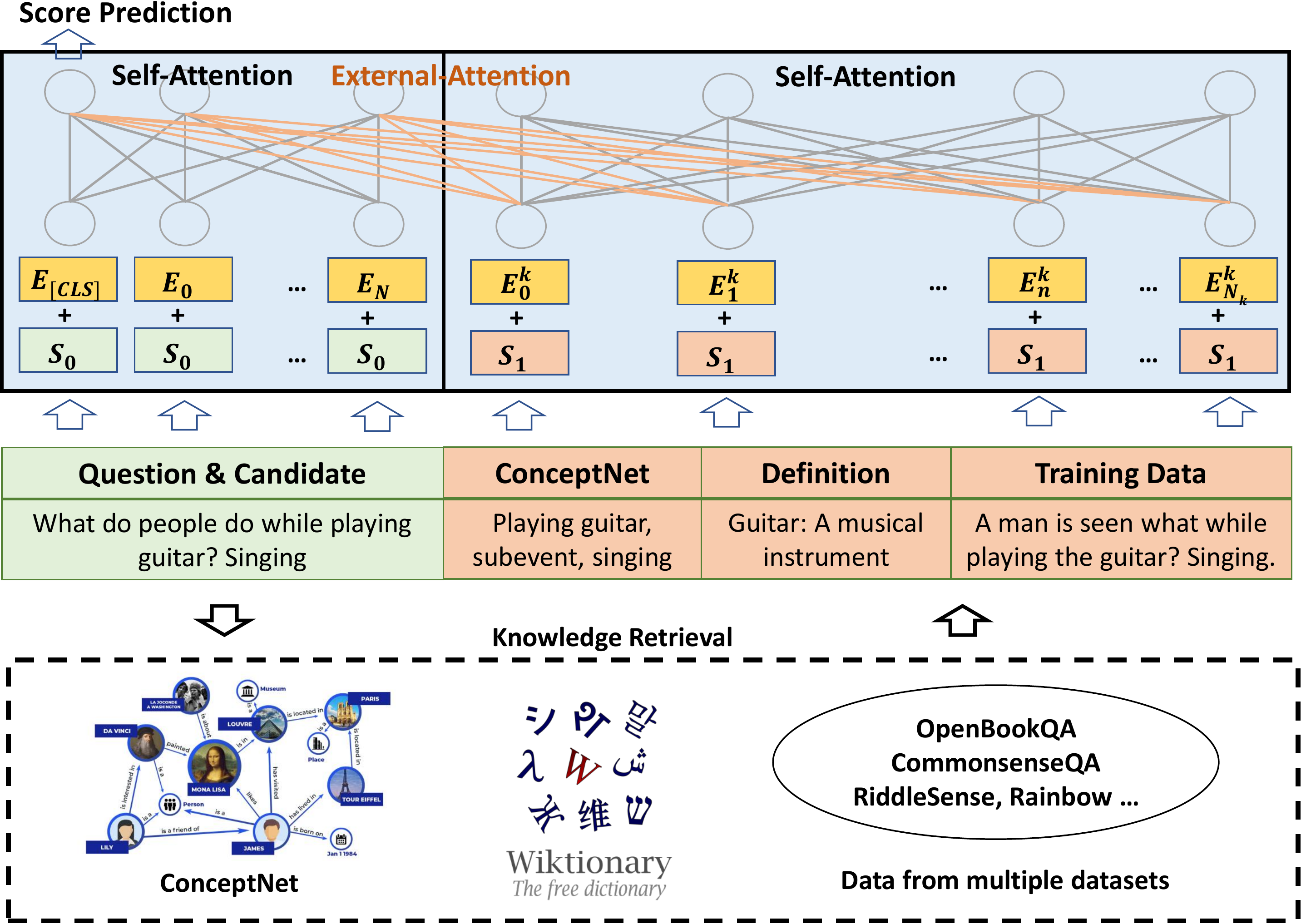}
	\caption{Our proposed method of Knowledgeable External Attention for commonsense Reasoning (KEAR). Related knowledge is retrieved from external sources, e.g., knowledge graph, dictionary and training data, using the input as the key and then integrated with the input. While additional external attention layers can be added to the Transformer blocks, we adopt text-level concatenation for external attention, incurring no structural change to the model architecture. 
		\label{fig:ea}}
\end{figure*}

While Transformer models process input by looking \textit{inward} via self-attention, we propose to make the model look \textit{outward} by providing it with related context and knowledge from various sources. We then let the model conduct self-attention on the input while also computing external attention to the knowledge (Figure~\ref{fig:ea}).
As the context and  knowledge can usually be stored in a non-parametric and symbolic way (e.g., plain text, knowledge graph and dictionary entries), 
even moderately-sized Transformer models can perform exceptionally well on NLP tasks.
This approach allows one to shrink the size of Transformer-based foundation models, which is critical to the accessibility and democratization of AI technology.
This approach is also analogous to the way humans conduct intelligence; we often resort to search engines, dictionaries, or information from other people to navigate the world.

Another benefit of external attention is that, as the related knowledge is stored outside of the model, practitioners can easily update the knowledge source to change the behavior of their models. For example, one could add or delete entries from a knowledge graph or rewrite certain paragraphs in Wikipedia. By explicitly representing knowledge, the decision process of the model becomes much more transparent and explainable. 

In this paper, we use the commonsense reasoning task CommonsenseQA~\citep{talmor2018commonsenseqa} as a case study in leveraging external attention to obtain and integrate information related to the input. Given a commonsense question and a choice, we retrieve knowledge from three external sources: a knowledge graph (ConceptNet), a dictionary (Wiktionary), and labeled training data (CommonsenseQA and 16 related QA datasets). The retrieved knowledge is directly appended to the input and sent to the language model with no revision to the underlying architecture. We show that with the proposed external attention, the accuracy of commonsense reasoning using a DeBERTa-xxlarge model~\citep{he2020deberta} can be significantly boosted from 83.8\% to 90.8\% on the dev set, while fine-tuned large-scale models like GPT-3 can only achieve 73.0\%. The ensembled version of our model, Knowledgeable External Attention for commonsense Reasoning (KEAR), reaches an accuracy of 93.4\% on the dev set and 89.4\% on the test set, surpassing human performance (88.9\%) for the first time~\citep{talmor2018commonsenseqa}.

The benefits of our approach extend beyond commonsense reasoning. First, the external attention dramatically reduces our system's dependence on large-scale models, i.e., achieving human parity with models up to 1.5B parameters. Second, the external information is obtained via computationally efficient methods, such as information retrieval and word matching, adding little computational cost to the main model. Third, the text-level concatenation of input and knowledge leads no change to the Transformer model, enabling existing systems to easily adopt this new external attention mechanism.

\section{Method}
We first describe our external attention framework in Sec \ref{sec:method_ext}. Next, we describe our external knowledge sources in Sec \ref{sec:method_knowledge}. Last, we present additional modeling techniques for improving commonsense reasoning in Sec \ref{sec:method_general}. 
\paragraph{Problem Formulation.} We focus on the multiple-choice question answering task in this paper, where the goal is to select the correct answer from a given list $c_1, c_2,..., c_n$ for a commonsense question $q$. The output of the model is a distribution $\mathcal{P}$ on $\{1,2,...,n\}$.
\subsection{External Attention\label{sec:method_ext}}
\paragraph{Self Attention.}
The majority of recent language models are based on the Transformer architecture \citep{transformer}. One of the most important components in Transformer is the self-attention mechanism, which can be formulated as
\begin{align}
	\bQ = \bH_l\bW_q, \bK &= \bH_l \bW_k, \bV = \bH_l \bW_v,\nonumber\\
	\bA = \frac{\bQ\bK^T}{\sqrt{d}}, &\;\bH_{l+1}	= \text{softmax}(\bA)V,
\end{align}
where $\bH_l\in \mathbb{R}^{N\times d}$ is the input hidden vectors to the $l$-th Transformer layer, $\bW_q, \bW_k, \bW_v\in \mathbb{R}^{d\times d}$ are projection matrices, $N$ is the input length and $d$ is the hidden vector's dimension. The inputs to the first Transformer layer are usually the embeddings of the tokenized input text, denoted as $\bH_0=\bX = [x_1, x_2, ..., x_N]$\footnote{We do not differentiate between tokens and their embeddings in the following discussion. Following previous work, we prepend a \texttt{[CLS]} token to the input.}. In the multi-choice question answering context, the input text is a concatenation of the question and a specific choice.

\paragraph{External Attention.} 
For commonsense question answering, the required information needed to answer the question is usually absent from the input. 
Thus, we need to integrate external knowledge into the model. 
In this work, we denote the extra knowledge in text format as $\bK = [x^K_1, x^K_2, ..., x^K_{N_k}]$. 
There are many ways to integrate the external knowledge into the model, such as using graph neural networks \citep{kagnet}. In this paper we simply concatenate the knowledge to the input text: 
$\bH_0 = [\bX; \bK] = [x_1, ..., x_N, x^K_1, ..., x^K_{N_k}]$. 
The advantage of this input-level integration is that the existing model architecture does not need to be modified.
Then, applying self-attention on $\bH_0$ can make the model freely reason between the knowledge text and the question/choices, therefore equipping the model with enhanced reasoning capacity.

\subsection{External Knowledge Sources\label{sec:method_knowledge}}
The knowledge to append to the input for external attention is crucial for getting the correct prediction. For commonsense reasoning, we collect three external knowledge sources to complement the input questions and choices.
\paragraph{Knowledge Graph.} Knowledge graphs (KG) contain curated facts that can help with commonsense reasoning which might not appear in a general corpus. We follow KCR\footnote{\url{https://github.com/jessionlin/csqa}}
to retrieve relevant relation triples in the ConceptNet graph~\citep{speer2017conceptnet}. Suppose the question entity is $e_q$ and the choice contains entity $e_c$\footnote{In CommonsenseQA dataset, both $e_q$ and $e_c$ are provided. Otherwise, we use entity linking to find related knowledge graph nodes to the input text (see ``Training Data'' part later in this section).}. If there is a direct edge $r$ from $e_q$ to $e_c$ in ConceptNet, we choose this triple $(e_q, r, e_c)$. Otherwise, we retrieve all the triples originating from $e_c$. We score each triple $j$ by the product of its confidence $w_j$ (provided by ConceptNet) and the defined relation type weight $t_{r_j}$: $s_j = w_j\cdot t_{r_j} = w_j \cdot \frac{N}{N_{r_j}}$, where $r_j$ is the relation type of $j$, $N$ is the total number of triples originating from $e_c$, $N_{r_j}$ is the number of triples with relation $r_j$ among these triples. We then choose the triple with highest weight. Finally, if the selected triple is $(e_1, r, e_2)$, we denote the knowledge from the KG as $\bK_{\text{KG}}(q,c) = [e_1\; r\; e_2]$.

\paragraph{Dictionary.} Although pre-trained language models are exposed to large-scale text data, the long tail distribution of words means that the quality of a word's representation is highly dependent on that word's frequency in the pre-training corpus.
Dictionaries, on the other hand, can provide accurate semantic explanation of words regardless of their frequency in datasets.
To help understand key concepts in the question and answer, we follow DEKCOR \citep{xu2021fusing} to use the Wiktionary definitions of the question and answer concepts as external knowledge. For every concept, we fetch the first (most frequent) definition from Wiktionary using its closest lexical match. Let $d_q$ be the definition text for $e_q$ and $d_c$ be the definition text for $e_c$, we denote the dictionary knowledge as $\bK_{\text{dict}}(q,c) = [e_q: d_q; e_c: d_c]$.

\paragraph{Training Data.} Although recent language models are giant in terms of the number of parameters, recent studies show that they cannot perfectly memorize all the details of their training data~\citep{reina}.

To tackle this challenge, we propose to retrieve relevant questions and answers from the training data as additional knowledge.
We use BM25 \citep{schutze2008introduction} to retrieve top $M$ relevant questions and answers from the training data. For each question $q$, we index the concatenation of the question text, the ground-truth choice $c^*$, ConceptNet triples and Wiktionary definitions: $[q; c^*; \bK_{\text{KG}}(q, c^*); \bK_{\text{dict}}(q, c^*)]$. For a new question $q'$ and a potential choice $c'$, we similarly build a query $[q'; c'; \bK_{\text{KG}}(q', c'); \bK_{\text{dict}}(q', c')]$ to retrieve most similar questions from the training set.
For each retrieved question from the training data, we drop the knowledge part and employ the retrieved question and the ground-truth answer as external knowledge. Suppose the retrieved questions and (correct) answers are $\{(q_1, c_1^*), (q_2, c_2^*), ..., (q_M, c_M^*)\}$, we denote the knowledge from training data as $\bK_{\text{train}} = [q_1\; c_1^*; q_2\; c_2^*; \cdots ; q_M\; c_M^*]$. During training, for each question $q$ we filter itself from the retrieved results to avoid data leakage. 

Different from~\citet{reina} where the retrieval questions are only obtained from the same dataset, we experiment with three sources of training data for retrieval: i) CSQA training data, ii) CSQA+OBQA+RiddleSense, a small collection of datasets focusing on ConceptNet knowledge, and iii) a pool of 17 datasets focusing on commonsense reasoning (we describe details of these 17 datasets in the Appendix). Since most datasets do not provide the question and choice entity $e_q, e_c$ for every question-choice pair, we use entity linking to find all entities $E_q = \{e_q^{(1)}, ..., e_q^{(n_q)}\}, E_c = \{e_c^{(1)}, ..., e_c^{(n_c)}\}$ appearing in the question and choice text respectively. We select the entity with the maximum length in $E_q$ and $E_c$ as the question and choice entity for Wiktionary definitions. For ConceptNet triples, we find edges between $E_q$ and $E_c$ and choose the one with the maximum total length.

Finally, we concatenate the retrieved knowledge from our three sources to form a final knowledge input: $\bK = [\bK_{\text{KG}}; \bK_{\text{dict}}; \bK_{\text{train}}]$. In practice, the semicolon is replaced by the separator token (e.g., \texttt{[SEP]}). We name our knowledge retrieval and integration technology as Knowledgeable External Attention for commonsense Reasoning (KEAR), shown in Figure~\ref{fig:ea}.

\subsection{General Methods to Improve Commonsense Reasoning\label{sec:method_general}}
Prior works have proposed other methods to improve general NLU performance, and it is therefore natural to wonder if these methods also work for commonsense reasoning.
Here, we explore two general methods for improving commonsense reasoning performance: i) using different text encoders and ii) virtual adversarial learning.

\paragraph{Text Encoders.} Previous methods for natural language understanding (NLU) 
have tried using BERT \citep{bert}, RoBERTa \citep{roberta}, ALBERT \citep{albert}, T5 \citep{t5}, ELECTRA \citep{electra} and DeBERTa \citep{he2020deberta} as the text encoder, achieving state-of-the-art performance on the GLUE benchmark \citep{wang2019glue}. Thus, we evaluate these models as encoders for the commonsense reasoning task.

\paragraph{Virtual Adversarial Training (VAT).} Previous works show that virtual adversarial training (VAT) can improve the performance for general NLU and question answering tasks \citep{jiang2020smart}.
In the multiple-choice commonsense reasoning task, the goal is to minimize the cross-entropy loss:
\begin{align}
	\min_{\theta} \mathbb{E}_{(x,y)\sim D}[\text{CE}(f(x;\theta),y)]
\end{align}
where $f$ produces the model prediction (distribution $\mathcal{P}$ on the choices), $\theta$ represents the model parameters, $y$ is the one-hot ground-truth answer vector, CE is cross-entropy, and $D$ is the empirical data distribution. VAT first finds the update $\delta$ that leads to the largest change in the predicted distribution, subject to a $L_2$-norm constraint. Then, a consistency regularization loss term is added to minimize the difference in the function's output when compared to the input variation $\delta$:

\begin{align}
	\min_{\theta}\;& \mathbb{E}_{(x,y)\sim D}[\text{CE}(f(x;\theta),y) + \label{eqn:vat}\\
	&\alpha\max_{\|\delta\|_2\leq \varepsilon} \text{CE}(f(x;\theta),f(x+\delta;\theta))],\nonumber
\end{align}
where $\alpha$ and $\varepsilon$ are hyperparameters.

\section{Experiments\label{sec:experiment}}
We present our empirical results in this section \footnote{Our source code is released at \url{https://github.com/microsoft/KEAR}. }. Each of our three external knowledge sources can boost the commonsense reasoning performance, and combining all the three techniques helps us reach the human parity on the CommonsenseQA benchmark. 

\subsection{Setup}
\paragraph{Data.} We focus on the CommonsenseQA (CSQA, \citeauthor{talmor2018commonsenseqa}, \citeyear{talmor2018commonsenseqa}) benchmark. CommonsenseQA is a widely used multiple-choice question answering dataset that requires commonsense knowledge. It contains 12k questions created using ConceptNet~\citep{speer2017conceptnet}. For an edge (subject, relation, object) in ConceptNet, \citet{talmor2018commonsenseqa} retrieve other object concepts with the same subject and relation as distractors for a question. A human worker is then asked to i) write a question containing the subject and with the object as the correct answer, ii) pick the most distractive answer from the retrieved concepts, and iii) write another distractor for the question. The final question contains five choices, with one correct choice, two random retrieved concepts, one human-picked concept, and one human-curated answer.

We present details of the 17 datasets that we use for training data retrieval in Table \ref{tab:datasets}. All the datasets are multiple-choice or classification datasets related to commonsense reasoning, and we include dataset details in the appendix. 
\begin{table}[htb]
	\centering
	\begin{tabular}{lccc}
		\thickhline
		\textbf{Dataset}& \textbf{Task} & \textbf{\#Train} & \textbf{\#Label} \\
		\hline
		$\alpha$NLI & NLI & 170k & 2\\
		SWAG & MC & 73.5k & 4\\
		RACE-Middle & MRC & 87.9k & 4\\
		CODAH & MC & 1672 & 4\\
		RiddleSense & MC & 3512 & 5\\
		SciTail & NLI & 23.6k & 2\\
		Com2Sense & MC & 808& 2\\
		AI2Science & MC & 1232 & 4\\
		WinoGrade & CoRef & 40.4k & 2\\
		CSQA & MC & 9741 & 5\\
		CSQA2.0 & CLF & 9264 & 2\\
		ASQ & MC & 8872 & 2\\
		OBQA & MC & 4960 & 4\\
		PhysicalIQA & MC & 16.1k & 2\\
		SocialIQA & MC & 33.4k & 3\\
		CosmosQA & MRC & 25.3k & 4\\
		HellaSWAG & NSP & 39.9k & 4\\
		\thickhline
	\end{tabular}
	\caption{\label{tab:datasets} The datasets used for training data retrieval. NLI stands for natural language inference, MC is multiple choice, MRC is machine reading comprehension, CLF is classification, NSP is next sentence prediction.}
\end{table}

\paragraph{Model Setup.} We feed the input text into a pretrained text encoder (e.g., DeBERTa) and take the representation $\bv\in \mathbb{R}^d$ of the \texttt{[CLS]} token, where $d$ is the dimension of the encoder. We set the segment id as 0 for the question and answer text, and 1 for the appended knowledge text. The position embedding simply runs from 1 to $l$, where $l$ is the total input length including the external knowledge. The embedding of \texttt{[CLS]} is projected to a scalar with learned weights and the final prediction is computed via a softmax
over all five choices for a question. We then minimize the cross entropy error during training.

\paragraph{Implementation Details.} We finetune the model using the AdamW optimizer \citep{loshchilov2017decoupled}. The batch size is set to 48 or smaller to fit the batch onto a single GPU. We train the model for 10 epochs and take the best result on the dev set. We choose the weight decay in $\{0, 0.01, 0.1\}$. The learning rate are chosen from $\{1e-5, 2e-5, 3e-6\}$ for all encoders except for DeBERTa; following the DeBERTa paper \citep{he2020deberta} we use a smaller learning rate, chosen from $\{4e-6, 6e-6, 9e-6\}$. We use the DeBERTa v2 model and choose from the pretrained model or model finetuned on MNLI. We also try out the recent DeBERTa V3 model \citep{he2021debertav3} which combines DeBERTa with adversarial pretraining.
For VAT, we choose the weight multiplier $\alpha \in \{0.1, 1.0, 10.0\}$ and set input variation norm $\varepsilon=1e-5$ (see Eqn. \ref{eqn:vat}). 
For retrieving from training data, we choose the data source with the best validation set performance from the three retrieval source datasets.
We set number of retrieved questions $M=10$.
We run each experiment with 3 different seeds and present results from the best run.

\subsection{Effects of Individual Components}
\begin{table}[t]
	\centering
	
	\begin{tabular}{lccc}
		\thickhline
		\textbf{Encoder}& \textbf{CSQA} & \textbf{MNLI} & \textbf{\#Para}\\
		\hline
		Fine-tuned GPT-3 & 73.0 & 82.1 & 175B\\
		RoBERTa-large & 76.7  & 90.2 & 355M\\
		ALBERT-xxlarge & 81.2 &  90.6& 235M\\
		ELECTRA-base & 75.0  & 88.8 & 110M\\
		ELECTRA-large & 81.3 & 90.9 & 335M\\
		DeBERTa-xlarge & 82.9 & 91.7 & 900M\\
		DeBERTa-xxlarge & 83.8 & 91.7 & 1.5B\\
		DeBERTaV3-large & 84.6 & 91.8 & 418M\\
		T5-11B & 83.5$^1$ & 91.3 & 11B\\
		\thickhline
	\end{tabular}
	\caption{CSQA dev set accuracy for various encoders. We append the accuracy on MNLI dataset (in-domain) for each encoder as a reference. MNLI scores are from the corresponding GitHub repositories. $^1$: from \citet{liu2021generated}.\label{tab:encoders}} 
\end{table}
\paragraph{General Methods.} As shown in Table \ref{tab:encoders}, there is a positive correlation between general performance on NLI tasks and commonsense reasoning abilities on CommonsenseQA. 
Notice that the fine-tuned GPT-3 model with 175 billion parameters could only achieve 73.0\% on the dev set of CommonsenseQA.
Based on these results, we choose ELECTRA-large and DeBERTa variants \citep{he2020deberta, he2021debertav3} as the encoders for subsequent experimentation. 

For virtual adversarial training, we find that VAT can improve commonsense reasoning accuracy for ELECTRA, improving the result from 81.3\% to 82.1\%. It does not show much improvement for DeBERTa models in our experiment. Therefore, we apply VAT to ELECTRA in subsequent experiments.

\begin{table}[th]
	\centering
	\begin{tabular}{lccc}
		\thickhline
		\textbf{Method}& \textbf{E-l+VAT} & \textbf{D-xxl} & \textbf{DV3-l} \\
		\hline
		Base & 82.1 & 83.8 & 84.6\\
		+ KG & 85.2 & 86.4 & 86.7\\
		+ Dictionary & 83.8 & 84.0 & 85.1\\
		+ Training data & 84.0 & 86.4 & 87.1 \\
		\thickhline
	\end{tabular}
	\caption{Applying external attention to different knowledge sources. E-l+VAT stands for ELECTRA-large with VAT, D-xxl stands for DeBERTa-xxlarge, DV3-l stands for DeBERTaV3-large.  \label{tab:external}}
\end{table}

\paragraph{Effect of External Attention.} As shown in Table \ref{tab:external}, all of the proposed knowledge sources bring gains in commonsense reasoning accuracy across all base encoder models. The dictionary, knowledge graph and training data bring 0.5\%, 2.1\%, and 2.5\% improvement, respectively, when DeBERTaV3-large \citep{he2021debertav3} is the base encoder model.

\begin{table}[htb!]
	\centering
	\begin{tabular}{lccc}
		\thickhline
		\textbf{Model}& \textbf{CSQA} & \textbf{3-Data} & \textbf{17-Data} \\
		\hline
		E-l+VAT & \textbf{84.0} & 82.9 & 82.8\\
		D-xxl & 86.2 & 86.1 & \textbf{86.4}\\
		DV3-l & 87.0 & \textbf{87.1} & \textbf{87.1}\\
		E-l+VAT + KG + Dict & \textbf{88.5} & 88.2 & 87.1\\
		D-xxl + KG + Dict & 89.8 & 90.5 & \textbf{90.8}\\
		DV3-l + KG + Dict & 91.0 & \textbf{91.2} & \textbf{91.2}\\    
		\thickhline
	\end{tabular}
	\caption{\label{tab:reina_source} Performance on CSQA dev set of model w.r.t source of training data retrieval. 
	E-l+VAT stands for ELECTRA-large with VAT, D-xxl stands for DeBERTa-xxlarge, DV3-l stands for DeBERTaV3-large.}
\end{table}
We find that the best training data retrieval source depends on the exact encoders and the techniques applied, and we show a comparison in Table \ref{tab:reina_source}. 
In general, the 17-dataset pool achieves the best performance for DeBERTa, but for ELECTRA retrieving from the CSQA training set alone can get the best performance. 
Table \ref{tab:external} and \ref{tab:reina_source} demonstrate the effectiveness of our proposed knowledge retrieval and concatenation methods.
\begin{table}[th]
	\centering
	\begin{tabular}{lc}
		\thickhline
		\textbf{Method}& \textbf{Dev Acc(\%)} \\
		\hline
		ELECTRA-large + VAT + KEAR & 88.7\\
		DeBERTa-xxlarge + KEAR & 90.8\\
		DeBERTaV3-large + KEAR & \textbf{91.2}\\
		Ensemble (39 models w/ KEAR) & \textbf{93.4} \\
		\thickhline
	\end{tabular}
	\caption{CSQA dev set results with different encoders and ensembles.\label{tab:dev}}
\end{table}

\begin{table}[htb!]
	\centering
	\begin{tabular}{lcc}
		\thickhline
		\textbf{Method} & Single & Ensemble\\
		\hline
		BERT+OMCS & 62.5 & - \\
		RoBERTa & 72.1 & 72.5 \\
		RoBERTa+KEDGN &-& 74.4\\
		ALBERT & -&76.5 \\ 
		RoBERTa+MHGRN & 75.4 & 76.5\\ 
		ALBERT + HGN & 77.3 & 80.0 \\
		T5 & 78.1 & - \\ 
		UnifiedQA & 79.1 & -\\
		ALBERT+KCR &79.5 &-\\	
		ALBERT + KD  & 80.3 & 80.9 \\
		ALBERT + SFR  & - & 81.8 \\
		DEKCOR  & 80.7 & 83.3 \\
		\hline
		Human & - & 88.9 \\
		\hline
		KEAR (ours) & \textbf{86.1} & \textbf{89.4}\\
		\thickhline
	\end{tabular}
	\caption{Results on test set from the leaderboard. The human performance is ensemble of 5 workers~\citep{talmor2018commonsenseqa}. \label{tab:parity} }
\end{table}

\begin{table*}[htbp]
	\centering
	\small
	\begin{tabular}{lp{13.5cm}}
		\toprule
		 Question &  What has a surface with many sides?  \\
		 Choices & \textbf{A) tetrahedron}, B) object,  C) geometry problem,  D) lake,  \textit{E) triangle} \\
		 KG & surface AtLocation \{tetrahedron, object, geometry problem, lake\}\\
		 Dictionary & surface: The overside or up-side of a flat object such as a table, or of a liquid.\\
		 & tetrahedron: A polyhedron with four faces.\\
		 Training Data & The four equal sides were all a smooth surface, he had carved and sanded a perfect what? tetrahedron.\\
		 \midrule
		 Question &  What is a treat that your dog will enjoy?   \\
Choices & A) salad, B) petted,  C) affection,  \textbf{D) bone},  \textit{E) lots of attention} \\
KG & dog Desires \{petted, affection, bone, lots of attention\}\\
Dictionary & dog: A mammal that has been domesticated for thousands of years.\\
& bone: A composite material making up the skeleton of most vertebrates.\\
Training Data & What do dogs like to eat? bones.\\		 
	 \bottomrule
\end{tabular}
\caption{Case study for the effect of external attention. We list two questions from the CSQA dataset with our retrieved knowledge (both retrieved training data questions are from CSQA). The correct answer is in \textbf{bold}, and our KEAR model selected the correct choice for both questions. We highlight the wrong choice that a DeBERTa 1.5B model chooses in \textit{italic}. \label{tab:case_study} }
\end{table*}

\subsection{Combining the Techniques}
Table \ref{tab:dev} shows the results of KEAR, which combines the best techniques in previous experiments, i.e., best encoders and external attention to all knowledge sources, to further boost the performance.
The best single model (DeBERTaV3-large + KEAR) achieves 91.2\% accuracy on the dev set.
To get the best performance, we train KEAR models with ELECTRA large, DeBERTa xlarge (900M), xxlarge (1.5B) and V3 large as encoders with 12 different seeds, resulting in 48 models in total. We rank the models by their dev set performance as $M_1, M_2, ..., M_{48}$. The ensemble prediction uses a majority vote on individual predictions. We picked the first $N$ models such that the dev set performance of ensembling $M_1, M_2,..., M_N$ is the best. We ended up with
39 models with 12 ELECTRA models, 12 DeBERTaV3 models, 11 DeBERTa-xxlarge models and 4 DeBERTa-xlarge models. Our ensemble model reaches 93.4\% accuracy on the dev set.
Table \ref{tab:parity} shows the official leaderboard result on the hidden test set. Our ensemble model exceeds the previously best DEKCOR model by over 6\% and exceeds the human performance (88.9\%) by 0.5\%.

\subsection{Case Study}

We present two examples from CSQA in Table \ref{tab:case_study} to illustrate how the model can reason between all retrieved knowledge sources to get the correct answer. For the first question, the knowledge graph helps rule out the wrong answer \textit{triangle} since it does not have a surface. The dictionary and training data attention further confirms that a tetrahedron has four sides/faces, which is the correct answer. For the second question, again knowledge graph rules out ``salad'' since a dog does not desire salads. The dictionary attention results suggest that bones are important for a dog, and training data attention suggests that bones are good food for a dog. This leads to the correct answer (bone). This suggests that all three knowledge sources are critical for getting the correct answer. Having access to all three knowledge sources makes it easier for model reasoning to get the correct answer.

\section{Related work}
\label{sec:rw}

Many previous works have proposed ways of incorporating external knowledge sources into Transformer architectures. For commonsense question answering, specialized knowledge graphs like ConceptNet \citep{speer2017conceptnet} and ATOMIC \citep{sap2019atomic} are the most popular choices for external knowledge \citep{chang2021incorporating,yao2022kformer,SONG202188}. \citet{kagnet} construct a scheme graph from concepts in the question and choices and uses an LSTM to reason on paths between question and choice concepts. 
\citet{yasunaga2021qa} construct a joint graph containing the QA context and KG, then use graph neural networks to reason over the two knowledge sources.  
\cite{10.1007/978-3-030-86362-3_22} proposes a KG-Transformer for using knowledge graphs in generative question answering. 

Another line of work explores less structured knowledge such as Wikipedia and dictionaries for commonsense reasoning~\citep{xu2021fusing,lv2020graph}. \citet{bhakthavatsalam2020genericskb} combine the knowledge from ConceptNet, WordNet, and other corpora to form 3.5M generic statements and show that this knowledge can help boost accuracy and explanation quality. \cite{mitra2020additional} compares several ways of incorporating external knowledge from a relevant corpus for commensense question answering.

Recently, there are approaches to generate facts from pretrained language models to complement missing facts in the external knowledge source. 
\citet{bosselut2019comet} 
finetune a pretrained model on ATOMIC for commonsense knowledge graph completion. \citet{liu2021generated} directly prompt the GPT-3 model \citep{gpt3} to get knowledge for reasoning. 

Beyond commonsense reasoning, external knowledge can also help boost performance on other language processing tasks like open domain question answering \citep{yu2021kg}, 
relation classification \citep{yu2020jaket} 
dialog response generation \citep{ghazvininejad2018knowledge}, 
conversational QA \citep{qin2019conversing}, 
multilingual NLU \citep{fang2021leveraging} 
and text generation \citep{yu2020survey}. Compared with prior work that uses extra modules (e.g., GNNs) or extra models (e.g., GPT-3), our external attention framework is extremely lightweight. It operates via a combination of non-parametric retrieval and text concatenation, which we show is highly effective, able to surpass human parity on the CommonsenseQA task.

\section{Conclusion}
\label{sec:conclusion}
We propose external attention as a lightweight framework for retrieving and integrating external knowledge for language understanding.
Compared with self-attention which benefits from ever-increasing model sizes, external attention can bring related information from external sources to supplement the input. We demonstrate that this strategy can lead to considerable gains in performance with little additional computational cost.
By leveraging knowledge from knowledge graphs, dictionaries, and training data, we show that our technology, KEAR, achieves human parity on the CommonsenseQA benchmark for the first time. For future work, we will apply the technique to other NLP tasks to improve language model performance with external knowledge.

\section*{Acknowledgement}
We thank the anonymous reviewers for their comments on our paper.
We thank Reid Pryzant for proof-reading the paper.

{
\bibliographystyle{named}
\bibliography{kear}
}


\end{document}